\documentclass[runningheads]{llncs_arxiv}
\usepackage[T1]{fontenc}
\usepackage{graphicx}
\usepackage{caption}
\usepackage{subcaption}
\usepackage{textcomp,mathcomp}
\usepackage{upgreek}

\usepackage{algorithmic}
\usepackage{amsmath,amssymb,amsfonts}
\usepackage{booktabs}
\usepackage{cite}
\usepackage{float}
\usepackage{glossaries}
\setacronymstyle{long-short}
\usepackage{makecell}
\usepackage{multirow}
\usepackage{tabularx}
\usepackage{xcolor}

\usepackage{soul}
\usepackage{pifont}

\newacronym{bsn}{BSN}{Body Sensory Network}
\newacronym{cuba}{CuBa}{Current-Based}
\newacronym{cuba-lif}{CuBa-LIF}{Current-Based LIF}
\newacronym{dl}{DL}{Deep Learning}
\newacronym{edp}{EDP}{Energy-Delay Product}
\newacronym{gmp}{GMP}{Gradual Magnitude Pruning}
\newacronym{har}{HAR}{Human Activity Recognition}
\newacronym{hpo}{HPO}{hyperparameter optimization}
\newacronym{imu}{IMU}{Inertial Measurement Unit}
\newacronym{iot}{IoT}{Internet of Things}
\newacronym{kpi}{KPI}{Key Performance Indicator}
\newacronym{lif}{LIF}{Leaky Integrate-and-Fire}
\newacronym{l2mu}{L$^2$MU}{LIF-based LMU}
\newacronym{lmu}{LMU}{Legendre Memory Unit}
\newacronym{ml}{ML}{Machine Learning}
\newacronym{mpu}{MPU}{microprocessor unit}
\newacronym{qat}{QAT}{Quantization-Aware Training}
\newacronym{rnn}{RNN}{Recurrent Neural Network}
\newacronym{rsnn}{RSNN}{Recurrent SNN}
\newacronym{snn}{SNN}{Spiking Neural Network}
\newacronym{wisdm}{WISDM}{Wireless Sensor Data Mining}

\usepackage{xcolor}
\usepackage{xurl}
\begin{document}
\title{Natively neuromorphic LMU architecture for encoding-free SNN-based HAR on commercial edge devices}
%
\titlerunning{Neuromorphic LMU for commercial edge devices}
%
\author{
Vittorio Fra\orcidID{0000-0001-9175-2838} \and 
Benedetto Leto\orcidID{0009-0007-8891-5051} \and 
Andrea Pignata\orcidID{0009-0001-4278-8348} \and 
Enrico Macii\orcidID{0000-0001-9046-5618} \and 
Gianvito Urgese\orcidID{0000-0003-2672-7593}
}
\authorrunning{V. Fra et al.}
%
\institute{Politecnico di Torino, Italy\\
\email{\{name.surname, andrea\_pignata\}@polito.it\\
benedetto.leto@studenti.polito.it}}
\maketitle              
\begin{abstract}
Neuromorphic models take inspiration from the human brain by adopting bio-plausible neuron models to build alternatives to traditional Machine Learning (ML) and Deep Learning (DL) solutions. 
The scarce availability of dedicated hardware able to actualize the emulation of brain-inspired computation, which is otherwise only simulated, yet still hinders the wide adoption of neuromorphic computing for edge devices and embedded systems.
With this premise, we adopt the perspective of neuromorphic computing for conventional hardware and we present the L$^2$MU, a natively neuromorphic Legendre Memory Unit (LMU) which entirely relies on Leaky Integrate-and-Fire (LIF) neurons.
Specifically, the original recurrent architecture of LMU has been redesigned by modelling every constituent element with neural populations made of LIF or Current-Based (CuBa) LIF neurons.
To couple neuromorphic computing and off-the-shelf edge devices, we equipped the L$^2$MU with an input module for the conversion of real values into spikes, which makes it an encoding-free implementation of a Recurrent Spiking Neural Network (RSNN) able to directly work with raw sensor signals on non-dedicated hardware.
As a use case to validate our network, we selected the task of Human Activity Recognition (HAR). We benchmarked our L$^2$MU on smartwatch signals from hand-oriented activities, deploying it on three different commercial edge devices in compressed versions too.
The reported results remark the possibility of considering neuromorphic models not only in an exclusive relationship with dedicated hardware but also as a suitable choice to work with common sensors and devices.

\keywords{Neuromorphic computing \and Edge devices \and IoT \and Encoding-free \and LIF neurons \and Legendre memory unit}
\end{abstract}

\section{Introduction}
Smart devices, wearable sensors and edge computing can be looked at as a fire triangle for efficient implementations of miniaturized, intelligent \glspl{bsn} in domains like healthcare, sport or working environment monitoring~\cite{dami2021predicting, capela2015feature, allahbakhshi2020using, ceolini2020hand-gesture, frank2019wearable,kulsoom2022review}.
The resulting variety of human-centered tasks is enclosed in the general definition of \gls{har}. Typically characterized by data collected with wearable devices through \glspl{imu}~\cite{lara2013survey, ferrari2021trends}, \gls{har} is in practice also a possible prototyping framework for \gls{ml} techniques, including \gls{dl} models, intended for time-varying signals~\cite{nweke2018deep, slim2019survey, demrozi2020human, khan2021survey}. 
In the \gls{dl} domain, \glspl{snn}~\cite{maass1997networks} offer an appealing approach to \gls{har} thanks to their unique capability of offering a twofold perspective: on the one hand, they are alternative to traditional methods in adopting the neuromorphic paradigm of driving inspiration from the human brain through bio-plausible neuron models; on the other hand, they can be deployed, besides the conventional hardware, on dedicated chips~\cite{Davies2018loihi, Mayr2019spinnaker, Orchard2021efficient, Muller-Cleve2022braille, Bos2023sub, Pedersen2023neuromorphic} designed to perform asynchronous and sparse computation.
Such dual character of \glspl{snn} translates into the need for diverse evaluation criteria. When we observe them through the lens of being alternative to non-spiking \gls{dl} solutions for synchronous digital devices, task-specific \glspl{kpi} have to be primarily accounted for; when neuromorphic hardware is targeted, energy and power are fundamental metrics to be included.
In both cases, however, the inherent hurdle they introduce to interface with data from the analog external world is the encoding step, needed to produce the spikes to be used as input for the network. 

In this work, we propose a solution to employ \glspl{snn} on non-dedicated edge devices to solve the \gls{har} task through an alternative approach to more traditional \gls{dl} methods. With our \gls{l2mu}, a natively neuromorphic implementation of the \gls{lmu} architecture~\cite{voelker2019legendre} where each block of the network is made of \gls{lif} neuron populations, we show a neuro-inspired model working with raw sensor data on conventional hardware. We use the \gls{wisdm} smartphone and smartwatch activity and biometrics dataset~\cite{weiss2019wisdm, weiss2019smartphone} as specific \gls{har} use case, and, thanks to the results achieved after model compression, we shine light on possible applications of the \gls{l2mu} in the domain of real-time \gls{iot} not only through still rare neuromorphic hardware but also with non-dedicated edge devices.

\section{Background}
\glspl{snn} and the \gls{har} task inherently have in common the fundamental role played by time, which carries significant information in both cases~\cite{roy2019towards,mekruksavanich2021deep}. As a result, neuromorphic computing can be successfully adopted by applying the former to solve the latter with both feedforward and recurrent architectures~\cite{Fra2022human}. 

\subsection{Human activity recognition}

\gls{har} is the classification task involving signals from human actions. Depending on the employed sensor, or sensor network, different classes of \gls{har} can be defined~\cite{gomaa2023perspective}. Among them, the ones relying on \gls{imu} data turn out to be attractive and promising thanks to the increasing diffusion of wearable devices suitable for easy and non-invasive monitoring of motion~\cite{ramanujam2021human}.

The \gls{wisdm} smartphone and smartwatch activity and biometrics dataset~\cite{weiss2019wisdm, weiss2019smartphone}, selected for this work, includes recordings derived from sensors worn by 51 subjects instructed to perform 18 different tasks lasting 3 minutes each.
The signals are acquired by capturing accelerometer and gyroscope readings with a sampling frequency of 20 Hz,  and the activities are grouped into three categories: i) non-hand-oriented activities, ii) hand-oriented activities related to general tasks, and iii) hand-oriented activities tied to eating actions.
Compared to its earlier version~\cite{kwapisz2011activity}, the dataset in this release provides a more balanced distribution of samples across the 18 activities.

\subsection{Legendre Memory Unit (LMU)}

The \gls{lmu} is a relevant \gls{rnn} architecture for classification tasks of time-based signals~\cite{voelker2020programming,Gupta2021comparing,Bartlett2021estimating} originally implemented in the Nengo framework~\cite{bekolay2014nengo}. We here summarize its main properties by reviewing the key definitions directly from~\cite{voelker2019legendre} for ease of reference.

A central attribute of the \gls{lmu} is the representation of a signal with temporal delay $u(t-\theta^{'})$ within a sliding time window of duration $\theta$ with $0 \leq \theta^{'} \leq \theta$. This process is achieved by means of a high-dimensional projection of the input signal $u(t)$ through the shifted Legendre polynomials. The element of degree $i$ in such polynomial basis is defined as in Eq.~\ref{equation1}:
\begin{equation}
\label{equation1}
P_i(x) = (-1)^i \sum_{j=0}^i \binom{i}{j} \binom{i+j}{j} (-x)^i
\end{equation}

\noindent
and it contributes to the input signal delaying as by Eq.~\ref{equation2}:
\begin{equation}
\label{equation2}
u(t-\theta^{'}) \approx  \sum_{i=0}^{d-1} P_i \left( \frac{\theta^{'}}{\theta} \right) m_i(t)
\end{equation}

\noindent
where the highest order ($d-1$) in the series expansion is related to the dimension $d$ of the state vector $\textbf{m}(t)$, defined by Eq.~\ref{equation3} as
\begin{equation}
\label{equation3}
\theta \dot{\textbf{m}}(t) = \textbf{Am}(t) + \textbf{B}u(t)
\end{equation}

\noindent
with \textbf{A} and \textbf{B} representing the ideal state-space matrices determined using the Padé approximants by means of Eq.~\ref{equation4} and Eq.~\ref{equation5} with $i, j \in [0, d - 1]$:
\begin{align}
    \textbf{A} &= [a]_{ij} \in \textbf{R}^{d\times d} &
    a_{ij} &= (2i + 1) \begin{cases}
                            -1 &  i < j \\
                            (-1)^{i-j+1} &  i \geq j
                        \end{cases} 
    \label{equation4}
    \\[0.3em]
    \textbf{B} &= [b]_i \in \textbf{R}^{d}  &
    b_i &= (2i + 1)(-1)^i 
    \label{equation5}
\end{align}

\subsection{Neuron models}

\glspl{snn}, also referred to as the third-generation neural networks, aim at emulating the asynchronous and sparse computation typical of the human brain~\cite{maass1997networks}. To achieve such a result, their elemental building block, i.e. the fundamental computational unit, is directly inspired by biological neurons, with typical definitions given by \gls{lif}-based models which offer a practically acceptable sacrifice of biological plausibility in favour of a low computational complexity~\cite{Izhikevich2007dynamical}.
For the \gls{l2mu}, working within the \texttt{snnTorch} framework\footnote{https://github.com/jeshraghian/snntorch/tree/master}, we selected two versions~\cite{Eshraghian2023training}:

\subsubsection{\texttt{Leaky} model}
The neuronal dynamics is described through the evolution in time of the membrane potential (Eq.~\ref{expression_lif}) and a spiking condition (Eq.~\ref{expression_spikes_emission}): 
\begin{equation}
    \centering
    \label{expression_lif}
    U_t = \beta \cdot U_{t-1} + W \cdot X_t - S_{t-1} \cdot \Theta
\end{equation}
\begin{equation}
    \label{expression_spikes_emission}
    S_t =
    \bigg \{
    \begin{array}{l}
    1, \quad if \quad U_t > \Theta \\
    0, \quad otherwise \\
    \end{array}
\end{equation}
where $\beta$ is the decay factor of the membrane potential, $t$ identifies the simulation time step, $W$ is the synaptic weights matrix, $X$ represents the input to the neuron and $\Theta$ represents the threshold voltage.

\subsubsection{\texttt{Synaptic} model} 
This \gls{cuba-lif} model includes an additional internal state $I_{syn,t} = \alpha \cdot I_{syn,t-1} +  W \cdot X_t$ to model a synaptic conductance, with the resulting neuronal dynamics described by Eq.~\ref{expression_cuba-lif} as:
\begin{equation}
    \centering
    \label{expression_cuba-lif}
    U_t = \beta \cdot U_{t-1} + I_{syn,t} - S_{t-1} \cdot \Theta
\end{equation}
Similarly to $\beta$, the decay factor $\alpha$ refers to an exponential decay.
The spiking condition is defined by Eq.~\ref{expression_spikes_emission} as for the \texttt{Leaky} model.

\section{Methodology}
Building a neuromorphic model means taking inspiration from discrete and sparse computation. Nonetheless, just as the human brain must account for real-world signals which can be considered continuous, so has to do every \gls{snn} to interface with traditional sensors. In this Section, we describe how we implemented our \gls{l2mu} to work with raw data and how we employed it to solve the \gls{har} task on commercial edge devices.

\subsection{Activities selection and segmentation}

Looking at final applications in real-time regimes, we applied a windowing procedure to the signals from both accelerometer and gyroscope in order to reduce the temporal length of individual samples. Specifically, we adopted a timescale of 2~s, consistent with the perception of continuous human-machine interaction~\cite{Miller1968response,Popovski2022perspective}, on the hand-oriented general activities. As a result, we obtained a subset of 7 classes with 36,201 non-overlapping samples made of 40 timestamps from the three axes of the accelerometer and the three axes of the gyroscope.
We then  performed a 60:20:20 partition to define training, validation, and test set respectively.

By selecting smartwatch signals from hand-oriented general activities, we focused on cross-domains actions and movements. We indeed aimed to prove, at least in a prototyping phase, the possibility to adopt our model in different contexts and for various applications dealing with wearable devices.

\begin{figure}[t]
    \centering
    \includegraphics[width=0.93\linewidth]{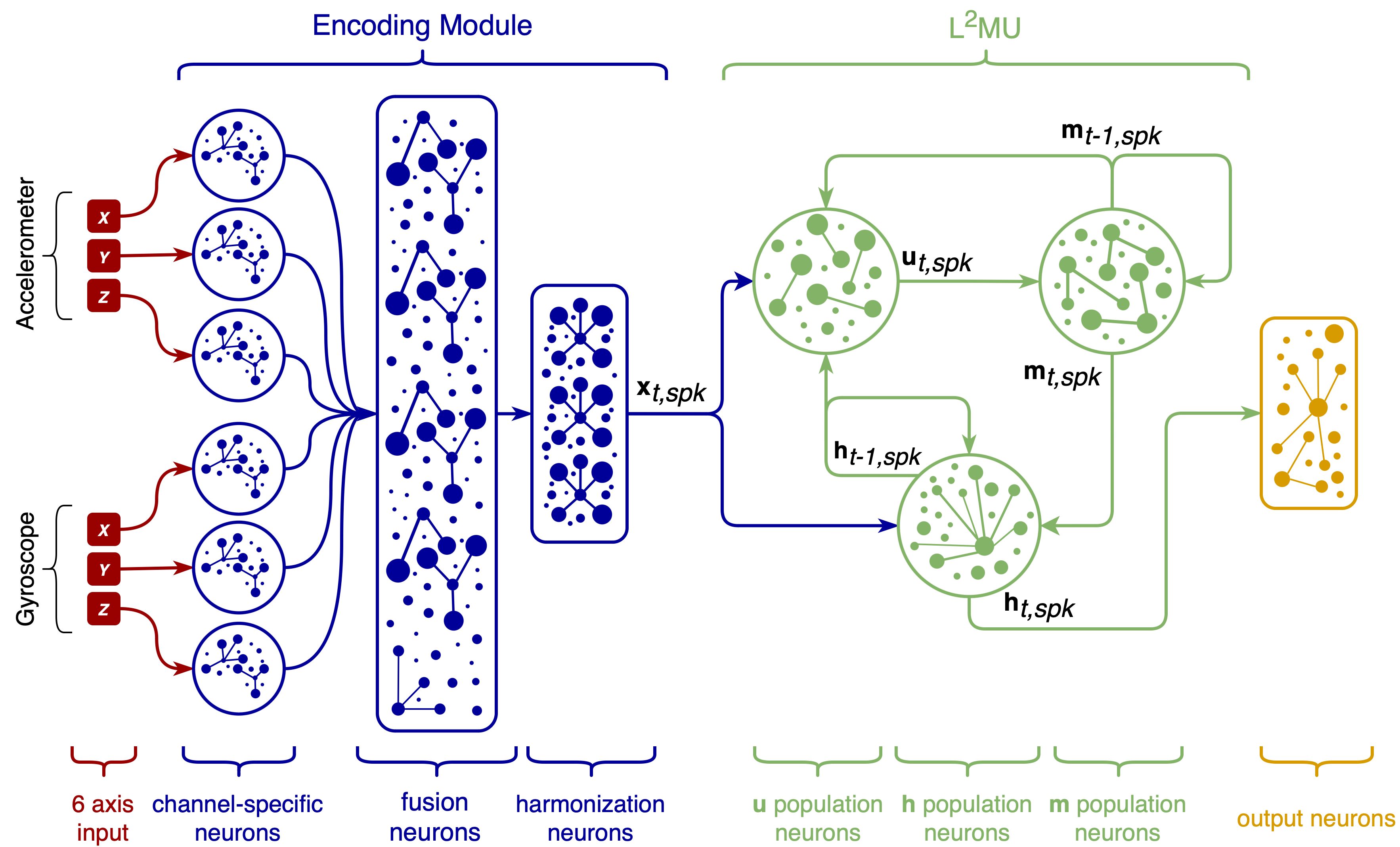}
    \caption{Schematic of the whole model. The 6-axis input is shown as six different channels in the leftmost part. Connected to them is the 3-layer, fully-connected, encoding module. It transmits spikes to the \gls{l2mu} which in turn feeds the output layer with the spikes from the hidden state (\textbf{h}) population.}
    \label{fig:network_architecture}
\end{figure}

\subsection{Encoding module}

An encoding module serves the purpose of converting the raw input signals into spike trains.
For our model, we adopted a three-layer, fully connected structure.
At the input side, signal expansion is performed by one LIF-based neural population specific for each input channel, which increases data dimensionality through the spiking activity. 
Such spikes are then propagated to the subsequent layer, where LIF-based neurons fuse the information coming from the channel-specific ones.
As final layer, LIF-based neurons are used to further harmonize the information contained in the spiking activity propagated across the encoding module.
The dimensions of the second and third layer are considered as hyperparameters of the whole model, and they undergo optimization.

\subsection{LIF-based LMU (L$^2$MU)}

Differently from other works in the \gls{snn} domain inspired by \gls{lmu}~\cite{Arsalan2023low,Gaurav2023reservoir,Liu2024lmuformer}, the natively neuromorphic redesign of \gls{lmu} we propose with the \gls{l2mu} relies on the conversion of every constituent element of the original architecture into neuron populations. In Fig.~\ref{fig:network_architecture}, the architecture is depicted, with the memory state, the hidden state and the projection on the Legendre polynomial basis represented by the \textbf{m} population, the \textbf{h} population and the \textbf{u} population respectively.
With this approach, we translated all the equations governing the interactions among the different components of the \gls{lmu} into neural activity.
The distinguishing inherent neuromorphic feature of the \gls{l2mu} hence is the neuro-inspired communication between the different building blocks, mediated by spikes generated as response to incoming currents.
Such conversion in terms of spiking activity is performed with two alternatives, relying on populations of neurons with \texttt{Leaky} or \texttt{Synaptic} model.

As explained in the original paper~\cite{voelker2019legendre}, the \gls{lmu} takes a value $\textbf{x}$ and produces a hidden state $\textbf{h}$ and a memory state $\textbf{m}$ to define the new representation $\textbf{u}$ of the input according to Eq.~\ref{equation_u}:
\begin{equation}
    \label{equation_u}
    \textbf{u}_{t} = \textbf{e}_{x}^\mathsf{T} \cdot \textbf{x}_t + \textbf{e}_{y}^\mathsf{T} \cdot \textbf{h}_{t-1} + \textbf{e}_{m}^\mathsf{T} \cdot \textbf{m}_{t-1}
\end{equation}
where $\textbf{e}_{x}$, $ \textbf{e}_{y}$ and $\textbf{e}_{m}$ are named as encoding vectors~\cite{voelker2019legendre}.
In the \gls{l2mu}, these steps are formulated with Eq.~\ref{equation_curr_u} relying on the spiking activity of the different populations of neurons: 
\begin{equation}
    \label{equation_curr_u}
    \textbf{u}_{t,curr} = \textbf{e}_{x}^\mathsf{T} \cdot \textbf{x}_{t,spk} + \textbf{e}_{y}^\mathsf{T} \cdot \textbf{h}_{t-1,spk} + \textbf{e}_{m}^\mathsf{T} \cdot \textbf{m}_{t-1,spk}
\end{equation}
with the corresponding spiking activity $\textbf{u}_{t,spk}$ produced through $\textbf{u}_{t,curr}$ as the neuronal input and $\textbf{u}_{t,mem} > \Theta^{u}$ as spiking condition coherently with Eq.~\ref{expression_spikes_emission}.

The new input representation is then written into the memory state, whose evolution in time can be discretized, starting from the original definition of Eq.~\ref{equation3}, as in Eq.~\ref{equation_m}:
\begin{equation}
    \label{equation_m}
    \textbf{m}_t = \overline{\textbf{A}} \cdot \textbf{m}_{t-1} + \overline{\textbf{B}} \cdot \textbf{u}_t
\end{equation}
where $\overline{\textbf{A}} = (\Delta t/ \theta) \cdot \textbf{A} + \textbf{I}$ and $\overline{\textbf{B}} = (\Delta t/ \theta) \cdot \textbf{B}$, with $\textbf{I}$ being the identity matrix and $\Delta t$ representing a time step within the window of length $\theta$. 
With similar arguments as for Eq.~\ref{equation_curr_u}, the spiking version of Eq.~\ref{equation_m} is defined by Eq.~\ref{equation_curr_m} as:
\begin{equation}
    \label{equation_curr_m}
    \textbf{m}_{t,curr} = \overline{\textbf{A}} \cdot \textbf{m}_{t-1,spk} + \overline{\textbf{B}} \cdot \textbf{u}_{t,spk}
\end{equation}
with the spiking activity $\textbf{m}_{t,spk}$ determined by such neuronal input $\textbf{m}_{t,curr}$ depending on the condition $\textbf{m}_{t,mem} > \Theta^{m}$ coherently with Eq.~\ref{expression_spikes_emission}.

Finally, the hidden state originally defined by Eq.~\ref{equation_h} as
\begin{equation}
    \label{equation_h}
    \textbf{h}_t = f(\textbf{W}_x \cdot \textbf{x}_t + \textbf{W}_{h} \cdot \textbf{h}_{t-1} + \textbf{W}_m \cdot \textbf{m}_t )
\end{equation}
is translated into Eq.~\ref{equation_curr_h}:
\begin{equation}
    \label{equation_curr_h}
    \textbf{h}_{t,curr} = \textbf{W}_x \cdot \textbf{x}_{t,spk} + \textbf{W}_{h} \cdot \textbf{h}_{t-1,spk} + \textbf{W}_m \cdot \textbf{m}_{t,spk}
\end{equation}
with the corresponding spiking activity $\textbf{h}_{t,spk}$ determined by $\textbf{h}_{t,curr}$ as neuronal input and the spiking condition $\textbf{h}_{t,mem} > \Theta^{h}$ coherently with Eq.~\ref{expression_spikes_emission}.

\subsection{Model optimization}

In order to identify the best configuration for our model to achieve as high as possible classification accuracy for the selected task, we carried out \gls{hpo} experiments.
Particularly, we employed the Neural Network Intelligence (NNI) toolkit\footnote{https://github.com/microsoft/nni} by designing a specific procedure and hyperparameter search space for each of the two neuron models accounted for.
We explored both architecture-related and neuron-related hyperparameters, thus performing a combined optimization aimed at identifying the best possible combination of neuron parameters ($\Theta$, $\alpha$ and $\beta$) and model structure.

All the optimization trials were performed by training the model for 300 epochs, and the validation accuracy was used as objective metrics. At the end of each trial, we evaluated the test accuracy, and the model achieving the highest result was selected as the best model of each \gls{hpo} experiment.
The following optimal dimensions were identified for the \texttt{Leaky}-based and the \texttt{Synaptic}-based \gls{l2mu} respectively: 30 and 30 channel-specific neurons per population, 170 and 180 fusion neurons, 10 and 10 harmonization neurons, 150 and 230 \textbf{u}~population neurons, 60 and 180 \textbf{h}~population neurons, 1050 and 1840 \textbf{m}~population neurons.

Following the \gls{hpo}, we retrained with different seed values the best models found, in order to acquire deeper information about the classification performance through a statistical analysis.

\subsection{Model compression}

To further prepare the \gls{l2mu}-based models for real-time \gls{iot} applications on conventional edge devices, pruning of the optimized models, with subsequent fine-tuning, has been investigated.
We compressed both the \texttt{Leaky}-based and the \texttt{Synaptic}-based \gls{l2mu} by means of the \texttt{sconce} Python library\footnote{https://github.com/satabios/sconce}. Specifically, starting from the pre-trained optimal model, we performed \gls{gmp} followed by fine-tuning for 300 epochs.
We adopted the same statistical approach as for the non-compressed models, using ten different seed values to perform the two steps.
As for the \gls{hpo} experiments, we selected as the best model for each neuron type the one providing the highest test accuracy.

\subsection{Deployment on hardware}

For the deployment of our encoding-free neuromorphic models on conventional edge devices, we identified three commercial boards embedding different ARM-based \glspl{mpu} oriented to edge applications. The first one is a STM32MP157F-DK2 by ST Microelectronics, which is based on the STM32MP157F \gls{mpu} with two ARM-A7 cores running at 800~MHz and a 32-bit architecture, and includes 500~MB of DDR3 RAM. The second one is a Raspberry Pi 3B+, with a Broadcom BCM2837B0 \gls{mpu} (four ARM-A53 cores, 1.4~GHz, 64-bit) and 1~GB of DDR2 RAM. The last one is a Raspberry Pi 4B, embedding a Broadcom BCM2711 (four ARM-A72 cores, 1.8~GHz, 64-bit) and 4~GB of DDR4 RAM. 
All the boards have a full Linux distribution: OpenSTLinux, based on the OpenEmbedded project, for the one from ST Microelectronics, and Raspberry Pi OS, based on Debian, for the Raspberry Pi ones. 

We deployed the compressed models through ONNX, performing on-edge inference with the ONNX Runtime Python library. The export of the model to ONNX was possible through \texttt{torch.onnx.export} thanks to the \texttt{PyTorch}-based foundation of \texttt{snnTorch}.

The boards ran the last versions of the operating system recommended by the manufacturers, with only the software needed for inference installed; and they were connected to a local network via WiFi. We monitored energy over USB-C connection while running the entire hardware for inference, and we measured RAM usage through the \texttt{psutil} Python library.

\section{Results}
The proposed encoding-free neuromorphic approach for \gls{har} has been evaluated from different perspectives. First, we statistically analysed the results from the training of the optimal models and their compressed versions; then, we carried out a more detailed comparison of the non-compressed models focusing on test accuracy. Finally, we investigated the inference phase for the deployed models.

\subsection{\texttt{Leaky} vs \texttt{Synaptic}}

\begin{figure}
    \centering
    \includegraphics[width=0.93\linewidth]{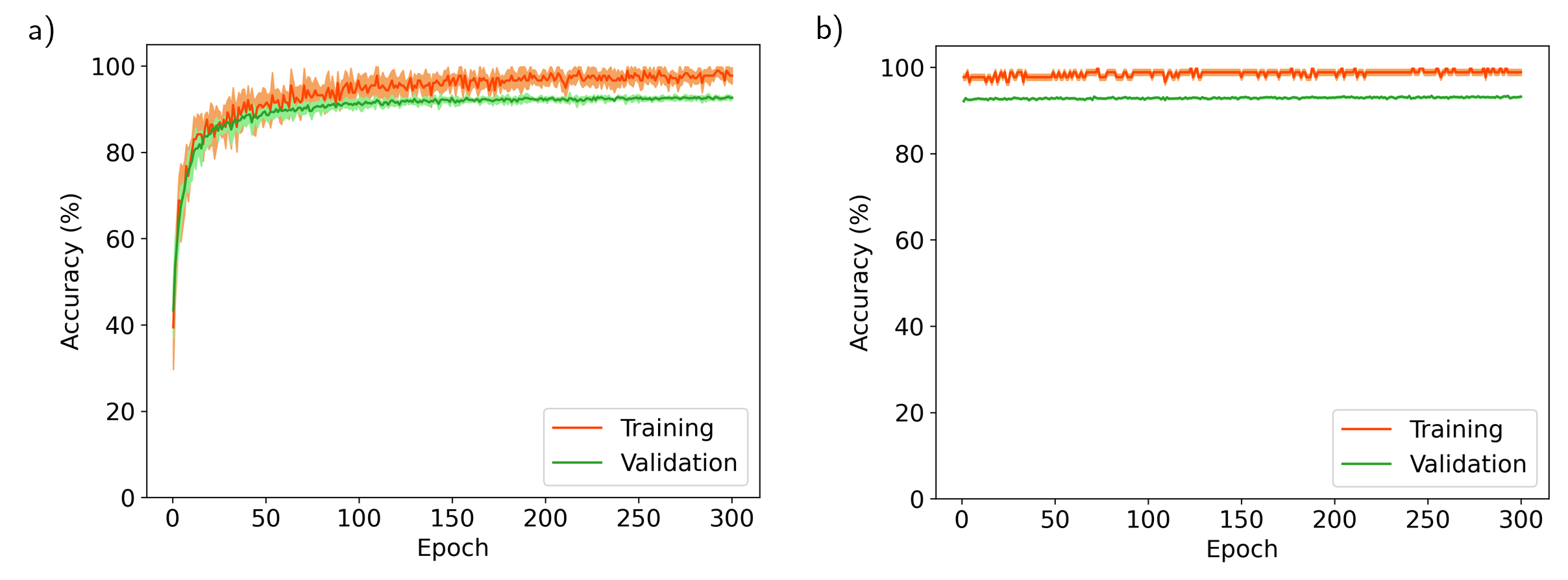}
    \caption{Learning curves from retraining of the best model with \texttt{Leaky} neurons. The accuracy for the optimal model identified through \gls{hpo} (a) is reported together with the results for the compressed model (b). Mean values and standard deviations are shown by solid line and shaded area respectively.}
    \label{fig:Leaky}
\end{figure}

\begin{figure}[t]
    \centering
    \includegraphics[width=0.93\linewidth]{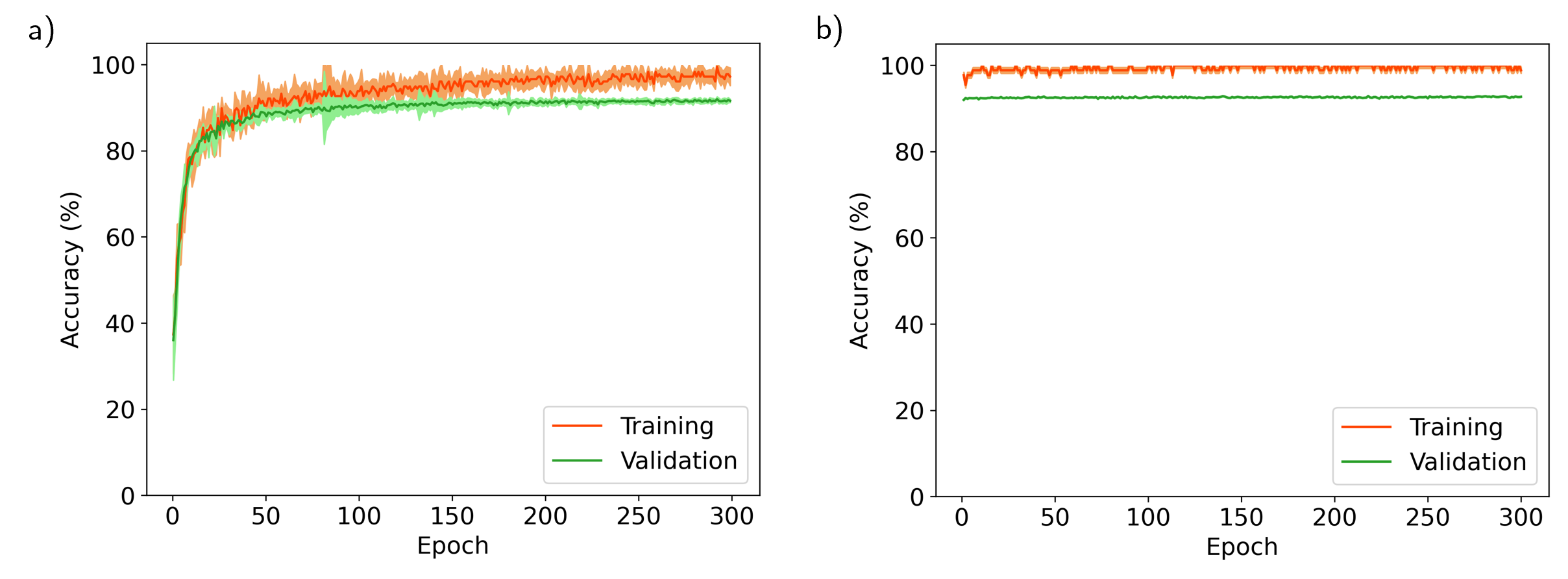}
    \caption{Learning curves from retraining of the best model with \texttt{Synaptic} neurons. The accuracy for the optimal model identified through \gls{hpo} (a) is reported together with the results for the compressed model (b). Mean values and standard deviations are shown by solid line and shaded area respectively.}
    \label{fig:Synaptic}
\end{figure}

\begin{figure}
    \centering
    \includegraphics[width=0.93\linewidth]{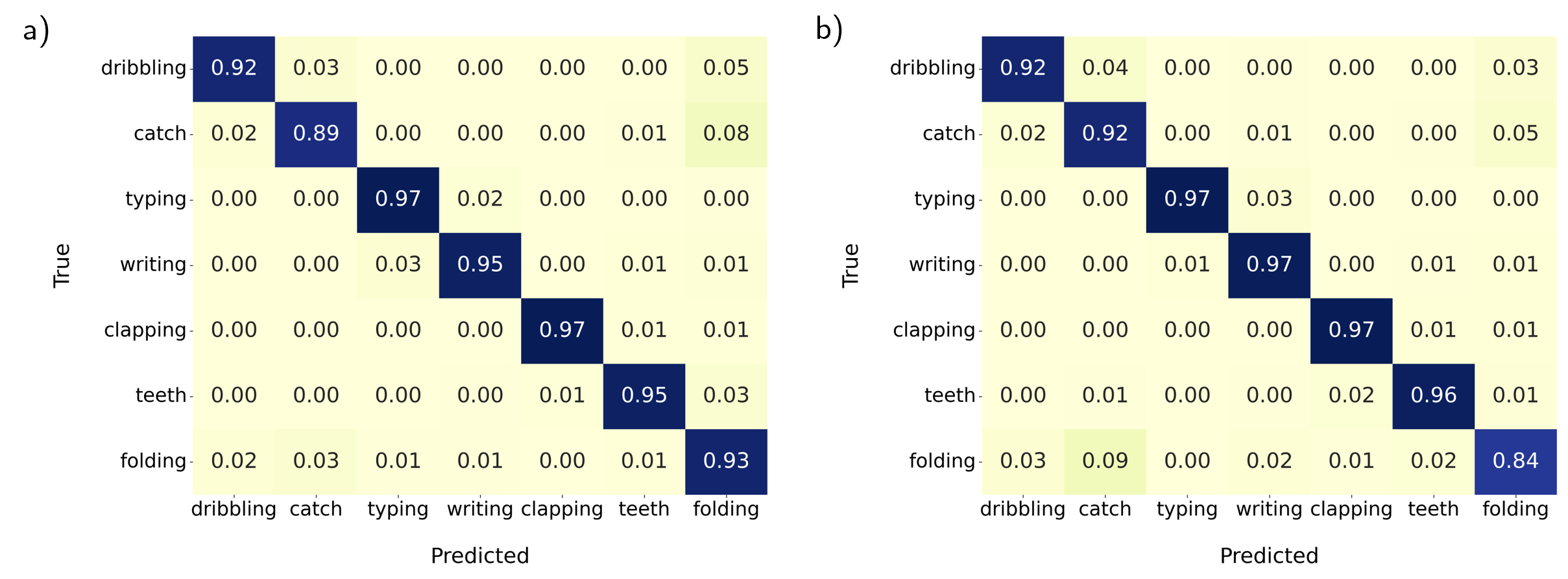}
    \caption{Illustrative confusion matrices for a detailed breakdown of the classification performance achieved by the optimized models after retraining with ten different seed values. With \texttt{Leaky} neurons (a), a maximum test accuracy of 94.14\% is achieved; while 93.48\% is the highest result obtained on test data with \texttt{Synaptic} neurons (b).}
    \label{fig:CMs}
\end{figure}

For each type of \gls{lif} neuron considered, we trained for 300 epochs, with validation at each of them, the optimal model obtained from \gls{hpo}, using ten different seed values.
For each seed, and for both the \texttt{Leaky}-based and the \texttt{Synaptic}-based model, we evaluated the test accuracy at the end of the training stage.
Fig.~\ref{fig:Leaky}a and Fig.~\ref{fig:Synaptic}a show the training and validation results, while test results are inspected in Fig.~\ref{fig:CMs} through confusion matrices.

Overall, \texttt{Leaky} neurons turned out to provide higher and more robust classification accuracy: the median value (93.29\%) is larger than the one achieved with \texttt{Synaptic} neurons (92.72\%), and the standard deviation is smaller (0.49\% instead of 0.61\%).
Compared to previous results on the same dataset~\cite{Fra2022human}, the \gls{l2mu} outperformed the standard \gls{lmu} with either neuron model. 

By analysing the disaggregated results through the classification accuracy per class reported in Fig.~\ref{fig:CMs}, it can be further highlighted that both the \texttt{Leaky} and the \texttt{Synaptic} neuron model provided high-level performance, overcoming chance level by more than 6 times and with only two classes below 90\%.

\subsubsection{Compressed models}

\begin{figure}
    \centering
    \includegraphics[width=0.93\linewidth]{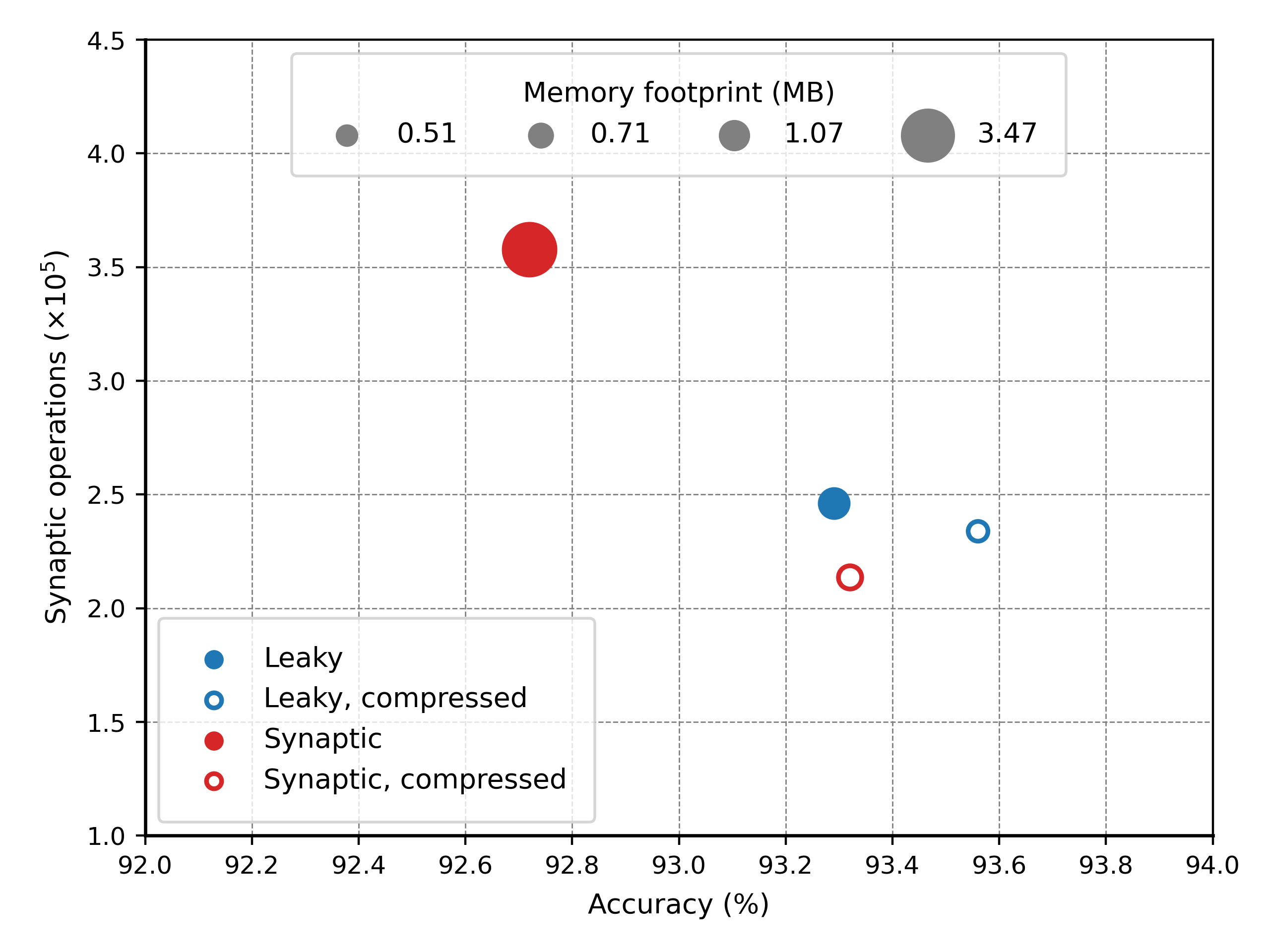}
    \caption{Comparison of the different models. Results from the median values of the statistical analyses are shown. With both \texttt{Leaky} and \texttt{Synaptic} neurons, an increase in test accuracy is achieved performing pruning and fine-tuning. On the contrary, the memory footprint and the number of synaptic operations per sample are reduced, for both neuron types, by the compressed models.}
    \label{fig:model_comparison}
\end{figure}

The learning curves resulting from pruning and subsequent fine-tuning for 300 epochs with different seed values are shown in Fig.~\ref{fig:Leaky}b and Fig.~\ref{fig:Synaptic}b for the \texttt{Leaky}-based and the \texttt{Synaptic}-based model respectively.
In Fig.~\ref{fig:model_comparison}, a summary is instead shown to compare different metrics for all the trained models, either with or without compression. 

With respect to the classification performance before compression, a slight improvement was achieved, with test accuracy equal to (93.56 $\pm$ 0.21)\% for \texttt{Leaky} neurons and (93.32 $\pm$ 0.23)\% for \texttt{Synaptic} neurons.

A remarkably beneficial impact of pruning was observed for the memory footprint. By evaluating the non-zero parameters after compression, the model size turned out to be reduced by 52.12\% and 79.52\% for the \texttt{Leaky}-based model and the \texttt{Synaptic}-based model respectively.

By leveraging the \texttt{NeuroBench} framework~\cite{Yik2024neurobench}, we also evaluated an additional neuro-inspired metrics: the number of effective synaptic operations per sample. Defined as the number of operations per sample performed on non-zero parameters, such quantity can be related to the computational effort in the neuromorphic perspective. It indeed provides a measure of how sparse and asynchronous operations could benefit energy saving. Compared to the corresponding non-compressed counterpart, a reduction of such operations by 4.99\% was achieved with \texttt{Leaky} neurons and by 40.25\% with \texttt{Synaptic} neurons.

\subsection{Deployed models}

\begin{table*}[t]
    \renewcommand{\arraystretch}{1.15}
    \centering
    \caption{Results achieved deploying the compressed models on commercial edge devices to solve the \gls{har} task on raw data}
    \label{table:hardware_LS}
    \begin{tabular}{{|>{\centering}m{1.5cm}|>{\centering}m{2.7cm}|>{\centering}m{1.4cm}|>{\centering}m{2.1cm}|>{\centering}m{1.9cm}|>{\centering\arraybackslash}m{1.8cm}|}}
        \hline
        { Neuron model} & { Device } & { Used RAM } & { Mean inference time } & { Mean energy per inference } & { Accuracy $^{(*)}$ } \\
        \hline
        \multirow{3.1}{*}{ \texttt{Leaky} } & { STM32MP1 } & { 65.7 MB } & { 0.13 s } & { 215.1 mJ } & \multirow{3.3}{*}{ 93.91\% } \\
        \cline{2-5}
        {  } & { Raspberry Pi 3B+ } & { 77.8 MB } & { 0.06 s } & { 268.8 mJ } & {  } \\
        \cline{2-5}
        {  } & { Raspberry Pi 4B } & { 77.4 MB } & { 0.03 s } & { 153.9 mJ } & {  } \\
        \hline
        \multirow{3.1}{*}{ \texttt{Synaptic} } & { STM32MP1 } & { 167.9 MB } & { 0.22 s } & { 383.4 mJ } & \multirow{3.3}{*}{ 93.84\% } \\
        \cline{2-5}
        {  } & { Raspberry Pi 3B+ } & { 187.5 MB } & { 0.15 s } & { 727.5 mJ } & {  } \\
        \cline{2-5}
        {  } & { Raspberry Pi 4B } & { 187.4 MB } & { 0.07 s } & { 348.9 mJ } & {  } \\
        \hline
    \end{tabular}
    \parbox[t]{0.99\textwidth}{$^{(*)}$ Evaluated on the whole test set}
\end{table*}

\begin{figure}[t]
     \centering
     \includegraphics[width=0.93\linewidth]{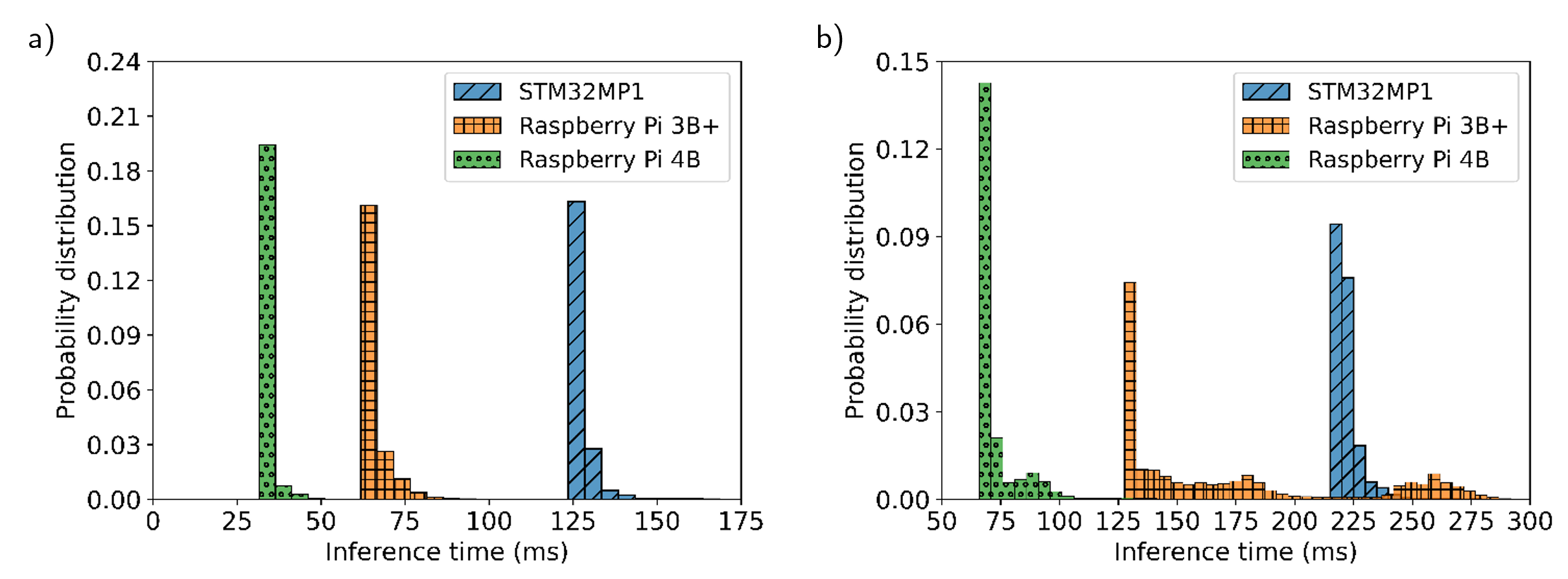}
   \caption{Inference time probability distribution for the deployed models with \texttt{Leaky} (a) and \texttt{Synaptic} (b) neurons. All the histograms have a bin width of 5~ms.}
   \label{fig:inference_time_probability_distribution}
\end{figure}

For the deployment on hardware, we considered the best compressed models after fine-tuning.
Table~\ref{table:hardware_LS} summarizes the results obtained on the three selected devices with single-sample inferences. For both \texttt{Leaky} and \texttt{Synaptic} neurons, RAM usage, mean inference time and mean energy per inference are reported, together with accuracy on the whole test set.
We also further investigated the latency of the deployed \glspl{l2mu} by running the models for 2~hours on the devices. Fig.~\ref{fig:inference_time_probability_distribution} show the inference time probability distributions for all the deployed models, providing a comparative analysis suitable for latency-based hardware selection.

The reported results show that both the types of \gls{lif} neuron can be deployed on traditional hardware with good results. Specifically, the \gls{l2mu} with the \texttt{Leaky} neuron model provided 93.91\% test accuracy with mean inference time of 0.13~s at most, and the \gls{l2mu} with \texttt{Synaptic} neurons reached a classification accuracy on test data of 93.84\% with inference time slightly higher: the mean values ranged from 0.07~s to 0.22~s.

Our \gls{l2mu} is hence capable, with both neuron types, of providing classification responses within a delay time consistent with an anthropocentric definition of real-time~\cite{Popovski2022perspective,Pelikan2023managing}, with values of \gls{edp} ranging, on average, from 4.6~mJ$\cdot$s with \texttt{Leaky} neurons on the Raspberry Pi 4B to 109.1~mJ$\cdot$s with \texttt{Synaptic} neurons on the Raspberry Pi 3B+.

\section{Conclusion}
In this work we have shown a successful application of neuromorphic models with off-the-shelf hardware through our \gls{l2mu}, a natively neuromorphic redefinition of the \gls{lmu} entirely based on populations of \gls{lif} neurons.
We built it in the \texttt{snnTorch} framework by transforming all the components of the \gls{lmu} into neural populations, and we equipped it with an encoding module to allow direct interface with traditional sensors collecting real-valued data.
By means of pruning and subsequent fine-tuning, we produced compressed versions of the proposed model, evaluating them on the \gls{har} task by adopting an encoding-free approach to perform classification on raw data.
Through deployment on three different commercial edge devices, we reported various metrics collected during on-edge inference.
By analyzing all of them, \texttt{Leaky} neurons turned out to offer the most attractive solution, providing the best results in classification accuracy, memory footprint and \gls{edp}.

With inference times always lower than 300~ms and classification accuracy above 93\%, our \gls{l2mu} demonstrated capability to operate in real-time regimes suitable for \gls{iot} applications; and the possibility of deploying it on commercial edge devices to work with an encoding-free strategy establishes a promising link between neuromorphic models and non-dedicated hardware.

\subsubsection{Acknowledgements}
This research is funded by the European Union - NextGenerationEU Project 3A-ITALY MICS (PE0000004, CUP E13C22001900001, Spoke 6) and the Fluently project with Grant Agreement No. 101058680.
We acknowledge a contribution from the Italian National Recovery and Resilience Plan (NRRP), M4C2, funded by the European Union – NextGenerationEU (Project IR0000011, CUP B51E22000150006, “EBRAINS-Italy”).

%
%
%
\bibliographystyle{splncs04}
\bibliography{bibliography}

\begin{thebibliography}{10}
\providecommand{\url}[1]{\texttt{#1}}
\providecommand{\urlprefix}{URL }
\providecommand{\doi}[1]{https://doi.org/#1}

\bibitem{allahbakhshi2020using}
Allahbakhshi, H., Conrow, L., Naimi, B., Weibel, R.: {Using accelerometer and GPS data for real-life physical activity type detection}. Sensors (Switzerland)  \textbf{20} (2020). \doi{10.3390/s20030588}

\bibitem{Arsalan2023low}
Arsalan, M., Santra, A., Issakov, V.: Low power radar-based air-writing system using genetic algorithm-assisted spiking legendre memory unit. In: 20th European Radar Conference (EuRAD) (2023)

\bibitem{Bartlett2021estimating}
Bartlett, M.E., Stewart, T.C., Thill, S.: Estimating levels of engagement for social human-robot interaction using legendre memory units. In: ACM/IEEE International Conference on Human-Robot Interaction (2021)

\bibitem{bekolay2014nengo}
Bekolay, T., Bergstra, J., Hunsberger, E., DeWolf, T., Stewart, T.C., Rasmussen, D., Choo, X., Voelker, A.R., Eliasmith, C.: {Nengo: A Python tool for building large-scale functional brain models}. Frontiers in Neuroinformatics  \textbf{7} (2014). \doi{10.3389/fninf.2013.00048}

\bibitem{Bos2023sub}
Bos, H., Muir, D.: Sub-mw neuromorphic snn audio processing applications with rockpool and xylo. In: Embedded Artificial Intelligence. River Publishers (2023)

\bibitem{capela2015feature}
Capela, N.A., Lemaire, E.D., Baddour, N.: {Feature Selection for Wearable Smartphone-Based Human Activity Recognition with Able bodied, Elderly, and Stroke Patients}. PLOS ONE  \textbf{10} (2015). \doi{10.1371/journal.pone.0124414}

\bibitem{ceolini2020hand-gesture}
Ceolini, E., Frenkel, C., Shrestha, S.B., Taverni, G., Khacef, L., Payvand, M., Donati, E.: {Hand-Gesture Recognition Based on EMG and Event-Based Camera Sensor Fusion: A Benchmark in Neuromorphic Computing}. Frontiers in Neuroscience  \textbf{14} (2020). \doi{10.3389/fnins.2020.00637}

\bibitem{dami2021predicting}
Dami, S., Yahaghizadeh, M.: {Predicting cardiovascular events with deep learning approach in the context of the internet of things}. Neural Computing and Applications  \textbf{33} (2021). \doi{10.1007/s00521-020-05542-x}

\bibitem{Davies2018loihi}
Davies, M., Srinivasa, N., Lin, T.H., Chinya, G., Cao, Y., Choday, S.H., Dimou, G., Joshi, P., Imam, N., Jain, S., Liao, Y., Lin, C.K., Lines, A., Liu, R., Mathaikutty, D., McCoy, S., Paul, A., Tse, J., Venkataramanan, G., Weng, Y.H., Wild, A., Yang, Y., Wang, H.: {Loihi: A Neuromorphic Manycore Processor with On-Chip Learning}. IEEE Micro  \textbf{38} (2018). \doi{10.1109/MM.2018.112130359}

\bibitem{demrozi2020human}
Demrozi, F., Pravadelli, G., Bihorac, A., Rashidi, P.: {Human Activity Recognition Using Inertial, Physiological and Environmental Sensors: A Comprehensive Survey}. IEEE Access  \textbf{8} (2020). \doi{10.1109/ACCESS.2020.3037715}

\bibitem{Eshraghian2023training}
Eshraghian, J.K., Ward, M., Neftci, E., Wang, X., Lenz, G., Dwivedi, G., Bennamoun, M., Jeong, D.S., Lu, W.D.: Training spiking neural networks using lessons from deep learning. arXiv preprint arXiv:2109.12894  (2024)

\bibitem{ferrari2021trends}
Ferrari, A., Micucci, D., Mobilio, M., Napoletano, P.: {Trends in human activity recognition using smartphones}. Journal of Reliable Intelligent Environments  \textbf{7} (2021). \doi{10.1007/s40860-021-00147-0}

\bibitem{Fra2022human}
Fra, V., Forno, E., Pignari, R., Stewart, T.C., Macii, E., Urgese, G.: {Human activity recognition: suitability of a neuromorphic approach for on-edge AIoT applications}. Neuromorphic Computing and Engineering  \textbf{2} (2022). \doi{10.1088/2634-4386/ac4c38}

\bibitem{frank2019wearable}
Frank, A.E., Kubota, A., Riek, L.D.: {Wearable activity recognition for robust human-robot teaming in safety-critical environments via hybrid neural networks}. In: 2019 IEEE/RSJ International Conference on Intelligent Robots and Systems (IROS) (2019). \doi{10.1109/IROS40897.2019.8968615}

\bibitem{Gaurav2023reservoir}
Gaurav, R., Stewart, T.C., Yi, Y.: Reservoir based spiking models for univariate time series classification. Frontiers in Computational Neuroscience  (2023)

\bibitem{gomaa2023perspective}
Gomaa, W., Khamis, M.A.: A perspective on human activity recognition from inertial motion data. Neural Computing and Applications  (2023)

\bibitem{Gupta2021comparing}
Gupta, G., Kshirsagar, M., Zhong, M., Gholami, S., Ferres, J.L.: Comparing recurrent convolutional neural networks for large scale bird species classification. Scientific reports  (2021)

\bibitem{Izhikevich2007dynamical}
Izhikevich, E.M.: Dynamical systems in neuroscience. MIT press (2007)

\bibitem{khan2021survey}
Khan, N.S., Ghani, M.S.: {A Survey of Deep Learning Based Models for Human Activity Recognition}. Wireless Personal Communications  (2021). \doi{10.1007/s11277-021-08525-w}

\bibitem{kulsoom2022review}
Kulsoom, F., Narejo, S., Mehmood, Z., Chaudhry, H.N., Butt, A., Bashir, A.K.: A review of machine learning-based human activity recognition for diverse applications. Neural Computing and Applications  \textbf{34} (2022)

\bibitem{kwapisz2011activity}
Kwapisz, J.R., Weiss, G.M., Moore, S.A.: {Activity recognition using cell phone accelerometers}. ACM SIGKDD Explorations Newsletter  \textbf{12} (2011). \doi{10.1145/1964897.1964918}

\bibitem{lara2013survey}
Lara, O.D., Labrador, M.A.: {A Survey on Human Activity Recognition using Wearable Sensors}. IEEE Communications Surveys {\&} Tutorials  \textbf{15} (2013). \doi{10.1109/SURV.2012.110112.00192}

\bibitem{Liu2024lmuformer}
Liu, Z., Datta, G., Li, A., Beerel, P.A.: Lmuformer: Low complexity yet powerful spiking model with legendre memory units. arXiv preprint arXiv:2402.04882  (2024)

\bibitem{maass1997networks}
Maass, W.: {Networks of spiking neurons: The third generation of neural network models}. Neural Networks  \textbf{10} (1997). \doi{10.1016/S0893-6080(97)00011-7}

\bibitem{Mayr2019spinnaker}
Mayr, C., Hoeppner, S., Furber, S.: Spinnaker 2: A 10 million core processor system for brain simulation and machine learning. arXiv preprint arXiv:1911.02385  (2019)

\bibitem{mekruksavanich2021deep}
Mekruksavanich, S., Jitpattanakul, A.: {Deep Convolutional Neural Network with RNNs for Complex Activity Recognition Using Wrist-Worn Wearable Sensor Data}. Electronics  \textbf{10} (2021). \doi{10.3390/electronics10141685}

\bibitem{Miller1968response}
Miller, R.B.: Response time in man-computer conversational transactions. In: Proceedings of the December 9-11, 1968, fall joint computer conference, part I (1968)

\bibitem{Muller-Cleve2022braille}
M{\"{u}}ller-Cleve, S.F., Fra, V., Khacef, L., Peque{\~{n}}o-Zurro, A., Klepatsch, D., Forno, E., Ivanovich, D.G., Rastogi, S., Urgese, G., Zenke, F., Bartolozzi, C.: {Braille letter reading: A benchmark for spatio-temporal pattern recognition on neuromorphic hardware}. Frontiers in Neuroscience  \textbf{16} (2022). \doi{10.3389/fnins.2022.951164}

\bibitem{nweke2018deep}
Nweke, H.F., Teh, Y.W., Al-garadi, M.A., Alo, U.R.: {Deep learning algorithms for human activity recognition using mobile and wearable sensor networks: State of the art and research challenges}. Expert Systems with Applications  \textbf{105} (2018). \doi{10.1016/j.eswa.2018.03.056}

\bibitem{Orchard2021efficient}
Orchard, G., Frady, E.P., Rubin, D.B.D., Sanborn, S., Shrestha, S.B., Sommer, F.T., Davies, M.: {Efficient Neuromorphic Signal Processing with Loihi 2}. In: IEEE Workshop on Signal Processing Systems (SiPS). vol. 2021-Octob (2021). \doi{10.1109/SiPS52927.2021.00053}

\bibitem{Pedersen2023neuromorphic}
Pedersen, J.E., Abreu, S., Jobst, M., Lenz, G., Fra, V., Bauer, F.C., Muir, D.R., Zhou, P., Vogginger, B., Heckel, K., et~al.: Neuromorphic intermediate representation: A unified instruction set for interoperable brain-inspired computing. arXiv preprint arXiv:2311.14641  (2023)

\bibitem{Pelikan2023managing}
Pelikan, H., Hofstetter, E.: Managing delays in human-robot interaction. ACM Transactions on Computer-Human Interaction  (2023)

\bibitem{Popovski2022perspective}
Popovski, P., Chiariotti, F., Huang, K., Kal{\o}r, A.E., Kountouris, M., Pappas, N., Soret, B.: A perspective on time toward wireless 6g. Proceedings of the IEEE  (2022)

\bibitem{ramanujam2021human}
Ramanujam, E., Perumal, T., Padmavathi, S.: {Human Activity Recognition with Smartphone and Wearable Sensors using Deep Learning Techniques: A Review}. IEEE Sensors Journal  \textbf{21} (2021). \doi{10.1109/JSEN.2021.3069927}

\bibitem{roy2019towards}
Roy, K., Jaiswal, A., Panda, P.: {Towards spike-based machine intelligence with neuromorphic computing}. Nature  \textbf{575} (2019). \doi{10.1038/s41586-019-1677-2}

\bibitem{slim2019survey}
Slim, S.O., Atia, A., M.A., M., M.Mostafa, M.S.: {Survey on Human Activity Recognition based on Acceleration Data}. International Journal of Advanced Computer Science and Applications  \textbf{10} (2019). \doi{10.14569/IJACSA.2019.0100311}

\bibitem{voelker2019legendre}
Voelker, A., Kaji{\'c}, I., Eliasmith, C.: Legendre memory units: Continuous-time representation in recurrent neural networks. Advances in neural information processing systems  \textbf{32} (2019)

\bibitem{voelker2020programming}
Voelker, A.R., Eliasmith, C.: Programming neuromorphics using the neural engineering framework. Handbook of Neuroengineering  (2020)

\bibitem{weiss2019wisdm}
Weiss, G.M.: {WISDM Smartphone and Smartwatch Activity and Biometrics Dataset}. UCI Machine Learning Repository: WISDM Smartphone and Smartwatch Activity and Biometrics Dataset Data Set  \textbf{7} (2019)

\bibitem{weiss2019smartphone}
Weiss, G.M., Yoneda, K., Hayajneh, T.: {Smartphone and Smartwatch-Based Biometrics Using Activities of Daily Living}. IEEE Access  \textbf{7} (2019). \doi{10.1109/ACCESS.2019.2940729}

\bibitem{Yik2024neurobench}
Yik, J., den Berghe, K.V., den Blanken, D., Bouhadjar, Y., Fabre, M., Hueber, P., Kleyko, D., Pacik-Nelson, N., Sun, P.S.V., Tang, G., Wang, S., Zhou, B., Ahmed, S.H., Joseph, G.V., Leto, B., Micheli, A., Mishra, A.K., Lenz, G., Sun, T., Ahmed, Z., Akl, M., Anderson, B., Andreou, A.G., Bartolozzi, C., Basu, A., Bogdan, P., Bohte, S., Buckley, S., Cauwenberghs, G., Chicca, E., Corradi, F., de~Croon, G., Danielescu, A., Daram, A., Davies, M., Demirag, Y., Eshraghian, J., Fischer, T., Forest, J., Fra, V., Furber, S., Furlong, P.M., Gilpin, W., Gilra, A., Gonzalez, H.A., Indiveri, G., Joshi, S., Karia, V., Khacef, L., Knight, J.C., Kriener, L., Kubendran, R., Kudithipudi, D., Liu, Y.H., Liu, S.C., Ma, H., Manohar, R., Margarit-Taulé, J.M., Mayr, C., Michmizos, K., Muir, D., Neftci, E., Nowotny, T., Ottati, F., Ozcelikkale, A., Panda, P., Park, J., Payvand, M., Pehle, C., Petrovici, M.A., Pierro, A., Posch, C., Renner, A., Sandamirskaya, Y., Schaefer, C.J., van Schaik, A., Schemmel, J., Schmidgall, S., Schuman,
  C., sun Seo, J., Sheik, S., Shrestha, S.B., Sifalakis, M., Sironi, A., Stewart, M., Stewart, K., Stewart, T.C., Stratmann, P., Timcheck, J., Tömen, N., Urgese, G., Verhelst, M., Vineyard, C.M., Vogginger, B., Yousefzadeh, A., Zohora, F.T., Frenkel, C., Reddi, V.J.: Neurobench: A framework for benchmarking neuromorphic computing algorithms and systems. arXiv preprint arXiv:2304.04640  (2024)

\end{thebibliography}





\end{document}